\documentclass[11pt]{article}

\usepackage[preprint]{acl}

\usepackage{times}
\usepackage{latexsym}
\usepackage{amsmath}
\usepackage{booktabs}
\usepackage{tabularx}
\usepackage{multirow}
\usepackage{colortbl}
\usepackage{booktabs}
\usepackage{multirow}
\usepackage{graphicx}
\usepackage{xcolor}
\usepackage{flafter}
\usepackage{placeins}
\usepackage{algorithm}
\usepackage{algpseudocode}
\definecolor{tablegroup}{RGB}{230,230,230}
\definecolor{tableblue}{RGB}{235,243,255}
\usepackage[T1]{fontenc}
\usepackage[utf8]{inputenc}

\usepackage{microtype}

\usepackage{inconsolata}

\usepackage{graphicx}

\title{StateRAG: Explicit Retrieval State with Checked Updates}

\author{
  \textbf{Miaohe Niu}\textsuperscript{1},
  \textbf{Xinyu Liu}\textsuperscript{1},
  \textbf{Runsong Zhao}\textsuperscript{1}, \\
  \textbf{Pengxiang Li}\textsuperscript{2,$*$,\dag}, 
  \textbf{Jingbo Zhu}\textsuperscript{1,3},
  \textbf{Tong Xiao}\textsuperscript{1,3,$*$} \\
  \textsuperscript{1}School of Computer Science and Engineering, Northeastern University, China \\
  \textsuperscript{2}Hong Kong Polytechnic University, China \\
  \textsuperscript{3}NiuTrans Research, Shenyang, China \\
  \texttt{\{niumiaohe,xiaotong\}@mails.neu.edu.cn}, \texttt{pengxiang.li@connect.polyu.hk}
}

\begin{document}
\maketitle

\let\thefootnote\relax\footnotetext{\textsuperscript{*}Corresponding authors.}

\begin{abstract}
Multi-step retrieval-augmented generation requires tracking evolving information needs, selected evidence, and unresolved gaps. 
However, free-form trajectories conflate this history, forcing each step to implicitly reconstruct the current state, while existing explicit-state methods remain vulnerable to inconsistent state overwrites. 
To address these limitations, we introduce \textbf{StateRAG}, a framework that coordinates a \textit{Planner}, \textit{Navigator}, \textit{Retriever}, and \textit{Verifier} through an explicit retrieval state with checked updates. 
% StateRAG dynamically tracks the plan, trace, evidence, and sufficiency, while a controller validates each proposed update against the accepted state before applying it—rejecting inconsistent proposals and guiding revisions within a query budget. 
These roles propose updates across four state fields: the plan, retrieval trace, evidence, and sufficiency assessment. 
A dedicated controller validates each proposal against the accepted state, rejecting inconsistent updates and guiding revisions within the query budget. 
A final \textit{Reader} then answers strictly from the validated evidence.
Ablation experiments on a shared graph backend show that StateRAG outperforms free-form trajectory and unverified explicit-state baselines on LongBench by 7.7 and 4.1 Macro F1 points, respectively. 
These gains consistently extend to document QA and flat passage retrieval. 
Furthermore, StateRAG surpasses the strongest external baselines by 5.3, 5.0, and 4.3 points on LongBench, QASPER, and DocVQA.
\end{abstract}

\section{Introduction}
\label{sec:introduction}

Retrieval-augmented generation (RAG) equips large language models with external knowledge to answer complex, knowledge-intensive questions~\cite{lewis2020retrieval}. 
Answering multi-hop or document-level questions typically requires iterative retrieval: as new evidence is gathered, the system must continually adapt its search strategy while tracking unresolved gaps and accumulated facts~\cite{trivedi2023interleaving,zhu2021adaptive}.
Consequently, effective multi-step retrieval fundamentally relies on maintaining an up-to-date, accurate state of what has been discovered, what remains missing, and what should be retrieved next.

Iterative and agentic approaches address this by interleaving reasoning with retrieval operations~\cite{yao2023react,trivedi2023interleaving,du2026arag}. 
However, standard systems record these operations in \textit{free-form trajectories}, where current plans, retrieved evidence, tool observations, and historical attempts are mixed into an unstructured execution trace.
This design introduces two limitations:
(1) \textbf{Implicit State Reconstruction}: Subsequent steps must repeatedly infer and reconstruct the current retrieval state from a noisy, ever-growing context history, making state tracking complex and error-prone.
%(2) \textbf{Lack of Update Integrity}:Newly generated plans or retrieved passages may be noisy, invalid, or inconsistent with previously accepted facts, yet free-form trajectories lack explicit mechanisms to validate or reject such proposals.
%While explicitly structuring state fields partially mitigates state reconstruction, it does not prevent invalid proposals from overwriting valid context.
(2) \textbf{Unchecked State Updates}: Newly generated plans or retrieved passages may be invalid or inconsistent with previously accepted facts, yet free-form trajectories lack explicit mechanisms to validate or reject such proposals. While explicitly structuring state fields~\cite{zhu2021adaptive,wu2024stateflow} partially mitigates state reconstruction, it does not prevent invalid proposals from overwriting valid context.

\begin{figure*}[!t]
    \centering
    \includegraphics[width=\textwidth]
        {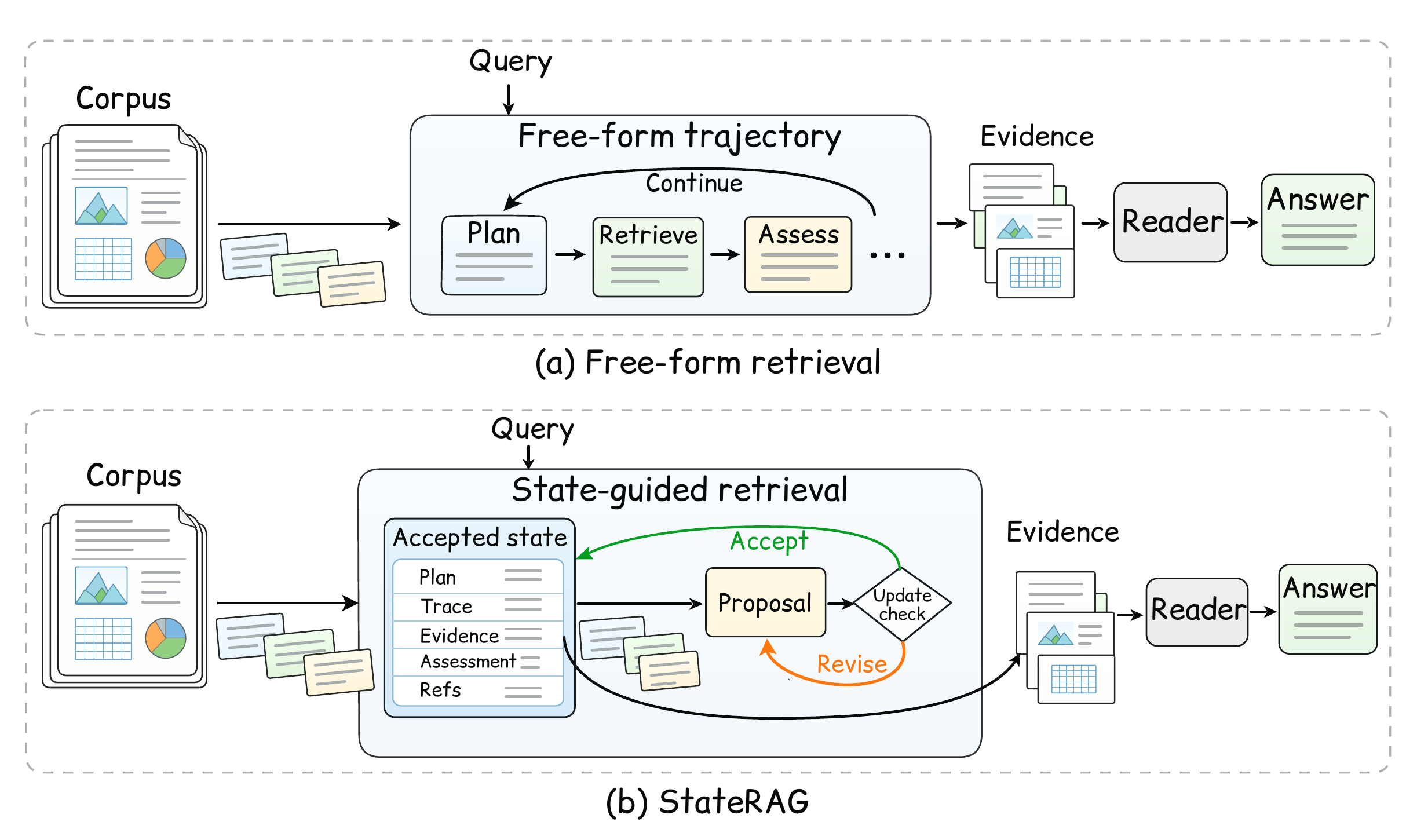}
    \caption{Retrieval-state management. (a) Free-form trajectories carry plans, observations, and assessments. (b) StateRAG checks proposed updates before acceptance and retains the accepted state after rejection.}
    \label{fig:state_concept}
\end{figure*}

In this paper, we propose to decouple the \textit{current retrieval state} from the \textit{historical execution trace} and treat state modifications as explicit, validated transitions (Figure~\ref{fig:state_concept}).
The state directly exposes four core fields to downstream steps: the current plan, retrieval trace, selected evidence, and sufficiency assessment. 
Crucially, proposed state updates are checked by a dedicated controller before acceptance. 
If a proposal is rejected, the system retains the accepted state and uses check feedback to guide iterative revision within a predefined query budget. 
This ensures that retrieval continuously advances from a grounded, validated state rather than being corrupted by erratic step proposals.

Building on this principle, we introduce \textbf{StateRAG}, a multi-agent framework that coordinates four functional roles—\textit{Planner}, \textit{Navigator}, \textit{Retriever}, and \textit{Verifier}—through a shared, typed retrieval state.
As illustrated in Figure~\ref{fig:state_method}, each role proposes updates to its corresponding state field, which are validated by the controller before execution proceeds. 
The system architecture integrates Typed Abstraction Memory (TAM) for structured candidate graph retrieval, the Multi-Agent Retrieval System (MARS) for role execution, and a Shared Memory Pool (SMP) for cross-query artifact reuse.

We evaluate StateRAG on multi-hop and long-document question answering benchmarks, including LongBench, QASPER, and DocVQA. 
StateRAG outperforms the strongest external baselines across all tasks, improving LongBench Macro F1, QASPER Answer F1, and DocVQA ANLS by 5.3, 5.0, and 4.3 points, respectively.
Extensive controlled experiments validate our key design choices:
holding the graph retrieval backend (TAM) fixed, replacing free-form trajectories with explicit state fields brings a 3.6-point gain on LongBench; introducing checked updates and state-retention mechanisms yields an additional 4.1-point improvement (7.7 points in total).
These gains consistently hold under a flat passage retrieval setting, confirming the generalizability of explicit state management with checked updates.
Our main contributions are summarized as follows:
\begin{itemize}
    \item We present a state-centric multi-step RAG framework where special roles coordinate via an explicit, typed state, disentangling current execution needs from noisy context histories.
    \item We design a checked-update mechanism that validates state change proposals, retains accepted states upon rejection, and leverages check feedback to guide budget-aware revisions.
    \item We demonstrate through comprehensive experiments that StateRAG achieves state-of-the-art performance across multiple benchmark datasets, with controlled ablations confirming the individual efficacy of explicit state fields and update checking.
\end{itemize}

\section{Related Work}
\label{sec:related_work}

\paragraph{Iterative retrieval and control.}
Multi-step retrieval methods coordinate information gathering with reasoning.
Self-Ask decomposes questions into follow-up questions~\cite{press2023measuring},
ReAct interleaves reasoning with actions and observations~\cite{yao2023react},
and IRCoT alternates retrieval with intermediate reasoning~\cite{trivedi2023interleaving}.
Retrieval decisions can also respond to generation confidence or query
complexity. FLARE retrieves using anticipated text when generation confidence
is low~\cite{jiang2023active}, while Adaptive-RAG selects among retrieval
strategies according to question complexity~\cite{jeong2024adaptive}.
Self-RAG learns reflection tokens for retrieval and critique~\cite{asai2024selfrag}.
CRAG evaluates retrieval quality to select subsequent retrieval and
evidence-processing actions~\cite{yan2024corrective}.
More recent work learns retrieval chains in CoRAG~\cite{wang2025chain},
exposes keyword search, semantic search, and chunk reading in
A-RAG~\cite{du2026arag}, and learns when to stop retrieval in
Stop-RAG~\cite{park2025stoprag}. These methods guide the choice and
sequencing of retrieval operations. More broadly, CRITIC uses feedback
from external tools to check and revise model outputs~\cite{gou2024critic}.
StateRAG focuses on how the information supporting retrieval is organized
into explicit fields and how candidate updates become part of the
accepted state.

\paragraph{Explicit state and workflows.}
Prior work makes execution structure and generation constraints explicit
at several levels. AISO represents its belief state using the question
and candidate evidence, and uses learned scores to select passages and
update the evidence set~\cite{zhu2021adaptive}.
ReWOO separates planning, tool execution, and answer synthesis through
plans and evidence variables~\cite{xu2023rewoo}.
StateFlow represents workflows with explicit states and transitions,
including verification and error handling~\cite{wu2024stateflow}.
RAG-Gym models information seeking as a decision process whose state
contains the question, reasoning history, and retrieved documents.
It improves action selection through process supervision and
critic-guided inference~\cite{xiong2026raggym}.
At the generation level, LMQL combines prompts with program control
flow and output constraints~\cite{beurerkellner2023lmql}, while PICARD
uses incremental parsing to reject inadmissible tokens during
decoding~\cite{scholak2021picard}.
StateRAG focuses on the content and update semantics of retrieval state.
It defines field meanings and update sources, checks candidate changes
before applying them, and retains the accepted state when a change is
rejected. Check feedback then guides revision within the query budget.

\paragraph{Structured retrieval.}
Document organization provides another source of retrieval structure.
DOS RAG preserves the original document order of retrieved
passages~\cite{laitenberger2025stronger}. RAPTOR recursively clusters and
summarizes passages into a tree for retrieval at multiple abstraction
levels~\cite{sarthi2024raptor}. GraphRAG builds an entity graph and
community summaries for query-focused synthesis~\cite{edge2024graphrag},
while HippoRAG2 integrates passages into a graph-based retrieval
memory~\cite{gutierrez2025hipporag2}. In StateRAG, TAM supplies a typed
directed acyclic graph for selecting retrieval regions and obtaining
evidence. Its outputs enter the shared state-update interface.
Replacing TAM with Flat passage retrieval provides a second backend
for evaluating the same state representation and update mechanisms.

\paragraph{Memory and artifact reuse.}
Memory systems preserve information beyond the immediate model context.
MemoRAG builds a global representation of long input contexts to guide
retrieval~\cite{qian2025memorag}. MemGPT manages information across memory
tiers~\cite{packer2023memgpt}, while Reflexion stores verbal feedback in
episodic memory to guide later trials~\cite{shinn2023reflexion}.
A-MEM organizes memories as linked notes and updates their relationships
as new information arrives~\cite{xu2025amem}. StateRAG connects stored
artifacts to retrieval state through references managed by SMP.
Role-private, query-scoped, and global storage distinguish access within
a query from reuse across queries. This separation supports within-query
state management and an additional analysis of sequential memory reuse.

\section{Method}
\label{sec:method}

\begin{figure*}[!t]
    \centering
    \includegraphics[width=\textwidth]
        {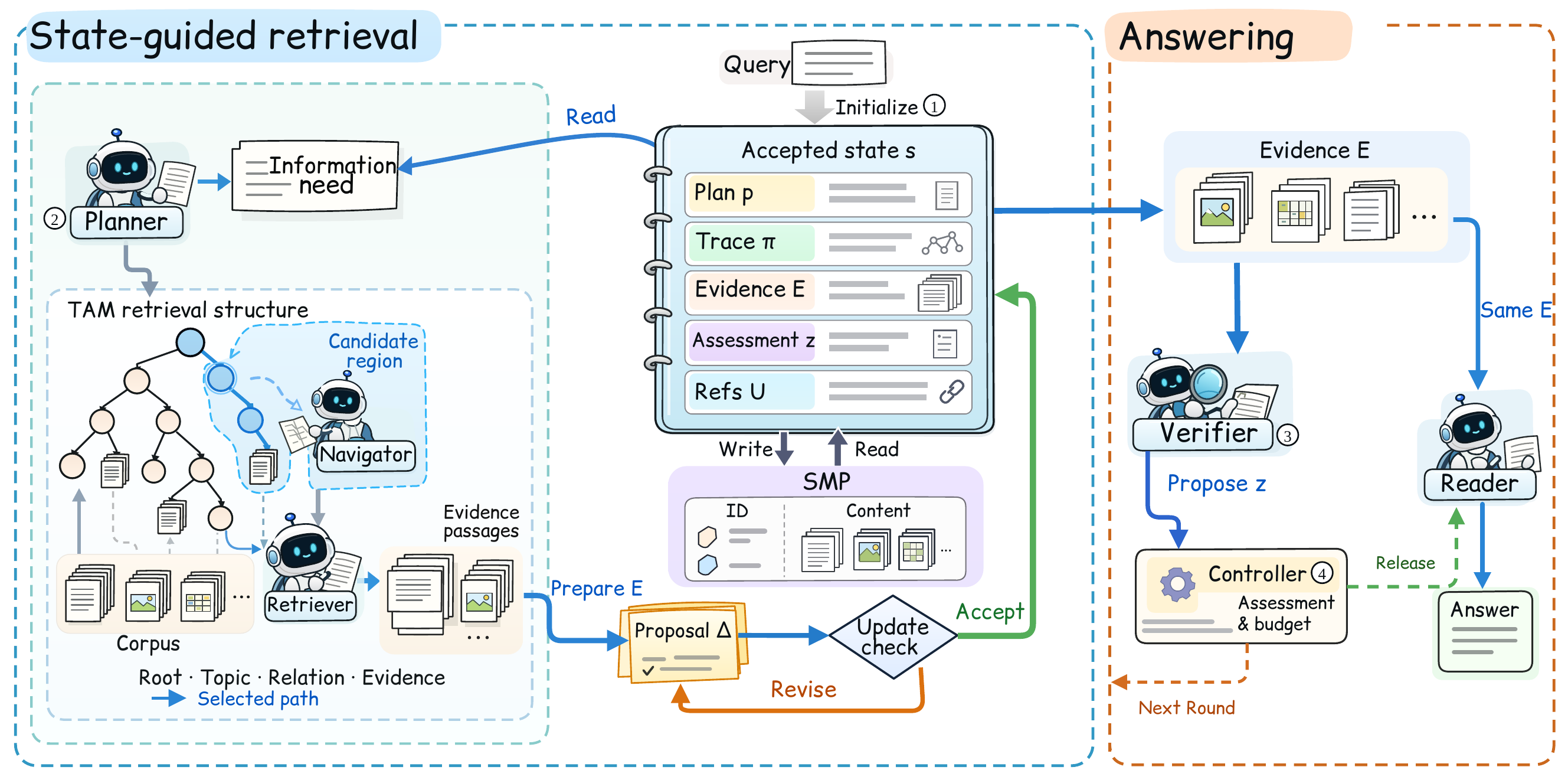}
    \caption{TAM Full's main iterative workflow. Stages 1--4 show initialization, planning and retrieval, evidence assessment, and action selection. Candidate updates are checked before acceptance. The controller selects actions using the assessment and remaining budget. On Release, the reader receives the assessed evidence unchanged.}
    \label{fig:state_method}
\end{figure*}

StateRAG maintains an explicit retrieval state throughout query processing.
Roles propose updates, which the controller checks before acceptance.
When a proposal is rejected, the controller retains the accepted
state. For recoverable proposals, check feedback guides revision
within the remaining budget.
Figure~\ref{fig:state_concept} contrasts StateRAG with free-form retrieval
trajectories, while Figure~\ref{fig:state_method} shows its retrieval and
state-update process.

\subsection{Retrieval State}
\label{sec:retrieval_state}

Consider a question comparing two people's birth years. After
finding one year, the system must retain its supporting evidence
and direct the next search toward the missing year. StateRAG
records this progress in five fields. For a query $q$, the
accepted state is
\begin{equation}
    s_t = (p_t, \pi_t, E_t, z_t, U_t).
    \label{eq:retrieval_state}
\end{equation}
Here, $t$ indexes update attempts rather than retrieval rounds.
We omit subscripts when referring to field definitions. The Shared Memory Pool (SMP) stores intermediate results, called artifacts, for reuse. The field $U$ holds references to these results rather than their contents.
Table~\ref{tab:state_fields} defines the fields and their update
sources.
\begin{table}[!htbp]
    \centering
    \small
    \setlength{\tabcolsep}{3pt}
    \renewcommand{\arraystretch}{1.08}
    \begin{tabularx}{\columnwidth}{
        @{}l >{\raggedright\arraybackslash}X l@{}
    }
        \toprule
        Field & Content & Update source \\
        \midrule
        Plan $p$
        & Current information need
        & Planner \\

        Retrieval trace $\pi$
        & Current path or retrieval record
        & Navigator \\

        Evidence $E$
        & Selected passages with provenance
        & Retriever \\

        Assessment $z$
        & Sufficiency verdict and justification
        & Verifier \\

        Artifact refs $U$
        & References to saved intermediate results
        & SMP \\
        \bottomrule
    \end{tabularx}
    \caption{Retrieval state fields and their update sources.}
    \label{tab:state_fields}
\end{table}

The trace $\pi$ records the current accepted retrieval step,
while earlier traces remain in the execution log. Evidence $E$
can retain passages from previous rounds with their source
identifiers. The assessment $z$ records evidence sufficiency and
any remaining gaps that guide the next plan. Each field has a
defined meaning, representation, and update source. Plans and
assessment justifications may contain natural language. The final
reader receives the query and selected evidence $E$, while the
other fields guide the preceding retrieval process.

\subsection{Retrieval Workflow}
\label{sec:retrieval_lifecycle}

Figure~\ref{fig:state_method} shows how StateRAG coordinates
retrieval through a shared state. In our main implementation,
Typed Abstraction Memory (TAM) organizes source material into
a graph. The Multi-Agent Retrieval System (MARS) coordinates
planning, graph navigation, evidence selection, and verification.
SMP stores intermediate results for reuse through references
in $U$. The final reader generates an answer from the selected
evidence.

An initial retrieval pass obtains candidate evidence.
Before iteration, \textsc{Bypass} sends nonempty initial evidence
directly to the reader if the complete accessible source collection
meets the source-unit limit and the untruncated evidence fits the
reader's evidence budget. The counting rules and limits are given
in Appendix~\ref{sec:additional_setup}.

Otherwise, the Planner identifies the next information need.
The Navigator specifies a retrieval path or query, and the
Retriever selects supporting passages. New passages are combined
with retained evidence, deduplicated, ranked, and truncated to
the reader's evidence budget. The resulting snapshot is proposed
as an update to $E$. Each role reads previously accepted updates,
and its own proposal is checked before changing the state.

Once the evidence update is accepted, the Verifier assesses the
snapshot and proposes an update to the assessment field $z$.
After this assessment update is accepted, the controller uses
the new verdict and the remaining budget to select the next
action. A \textsc{Pass} verdict triggers \textsc{Release}, which
sends the assessed snapshot unchanged to the reader.
A \textsc{Fail} verdict triggers \textsc{Revise} when the round
and call budgets permit further retrieval. The assessment
justification guides the next plan. Any revised evidence snapshot
is assessed again before the next decision.

When the remaining budget cannot support further retrieval or
revision, \textsc{Fallback} selects available accepted evidence
within the reader's budget. Execution ends without a reader call
if the required inputs cannot be formed. Across \textsc{Bypass},
\textsc{Release}, and \textsc{Fallback}, the final reader is called
at most once per query.

\subsection{Checked Updates and Revision}
\label{sec:checked_updates}

A role proposes an update $\Delta_t$ containing a target field,
a candidate value, and the producing role. The controller checks
the update source and field requirements. For example, a plan
must specify a retrieval objective, a TAM path must follow
permitted edges and node types, and selected evidence must retain
its source identifiers. The update takes effect as
\begin{equation}
    s_{t+1} =
    \begin{cases}
        \operatorname{Apply}(s_t,\Delta_t),
        & \operatorname{Check}(s_t,\Delta_t)=1,\\
        s_t,
        & \operatorname{Check}(s_t,\Delta_t)=0.
    \end{cases}
    \label{eq:checked_update}
\end{equation}

Here, $\operatorname{Apply}$ writes the candidate value to its
designated field. When a proposal is rejected, the controller
retains the accepted state. For recoverable proposals, feedback
from the update check guides revision within the remaining budget.
Revised proposals undergo the same checks, and revision calls
count toward the query budget.

The update check determines whether a proposal can enter the
state. The Verifier assesses whether the evidence is sufficient
to answer the query. A \textsc{Fail} assessment can therefore be
accepted as an update. It records the evidence gap and guides
further retrieval.

\subsection{Instantiation}
\label{sec:instantiation}

We refer to the complete implementation with TAM, MARS, and SMP
as TAM Full in the experiments.

\paragraph{TAM.}
TAM organizes document regions under relation and topic summaries
to guide retrieval. Construction begins with overlapping document
regions. Each region becomes an Evidence node retaining its source
identifier and offsets. Cosine-based agglomerative grouping forms
Relation groups from Evidence nodes and Topic groups from Relation
nodes. At each level, a group is represented by the normalized
mean of its members' embeddings. These representations are fixed
during cross-group assignment. A node can acquire a second parent
when its cosine similarity to another group's representation meets
the threshold and the node types and levels are compatible.
This adds an edge to the same node ID and preserves its original
parent. Each node has at most two parent groups.

After assignment at each level, an auxiliary language model
summarizes each group's final members. Relation summaries describe
relationships among evidence regions, and Topic summaries describe
shared subjects among Relation nodes. A root connects the Topic
nodes. Edges run only from a parent level to the adjacent child
level, yielding a directed acyclic graph. The Navigator scores
admissible path extensions using the query, plan, and node
abstractions. Its accepted root-to-Evidence path identifies the
region from which the Retriever selects passages for $E$. Appendix~\ref{app:tam_details} provides construction
and navigation parameters.

\paragraph{MARS.}
MARS runs the Planner, Navigator, Retriever, and Verifier in that
order. Role-specific prompts specify each role's task, inputs, and output format (Appendix~\ref{app:role_prompts}). The roles submit updates to $p$, $\pi$, $E$,
and $z$, respectively, using the state containing updates accepted
before their execution. The Navigator combines the current plan
with retrieval requests, results, and metadata supplied by the
controller and backend adapter to construct its proposal.
The Retriever uses accepted retrieval choices to select evidence,
which the Verifier then assesses against the query.

\paragraph{SMP.}
SMP stores document-processing outputs, generated abstractions, and retrieval records for reuse
(Appendix~\ref{app:smp_details}).
The field $E$ contains passages selected for assessment and
answering, while $U$ holds references to stored results.
Access depends on the memory scope and requesting role.
Query-scoped memory $M_t$ supports sharing among roles within
a query. Role-private memory $M_a$ serves its owning role
within that query. Both are initialized for each query.
Global memory $M_g$ retains results written during query
processing, allowing later queries to reuse them when the
store persists across a sequence. The sequential-memory
protocol is described in Appendix~\ref{sec:additional_setup}.

\paragraph{Flat control variant.}
For the controlled experiments, we replace TAM traversal with
passage retrieval. The Navigator submits a query and collects
ranked hits. Its proposed $\pi$ records the backend, round, index,
query text, requested top-$k$, and each hit's passage ID, rank,
and score. Once the controller accepts this record, the Retriever
fetches the passages by ID. They enter the shared evidence-update
and assessment stages.

% Requires amsmath.
% If the Experiments section already exists, insert from
% \subsection{Experimental Setup} onward.

\section{Experiments}
\label{sec:experiments}

\subsection{Experimental Setup}
\label{sec:experimental_setup}

We evaluate three tasks from LongBench~\cite{bai2024longbench},
HotpotQA~\cite{yang2018hotpotqa},
2WikiMQA~\cite{ho2020constructing}, and
MuSiQue~\cite{trivedi2022musique}, with 200 queries each.
We also evaluate the 1,451-query QASPER test set~\cite{dasigi2021qasper}
and 2,000 fixed DocVQA validation queries~\cite{mathew2021docvqa}.
For LongBench, we report EM and F1 and their unweighted means across
tasks, using Macro F1 for the main analysis. QASPER uses Answer F1
and DocVQA uses ANLS. Quality scores use a 0--100 scale.
All methods receive the same queries and source materials, with
cached image-to-text outputs shared on DocVQA.

We compare Free-form, Structured, and Full on both TAM and Flat.
Free-form uses an append-only text trajectory with role labels,
while Structured uses explicit state fields. Full adds update
checks, retention of accepted state, and feedback-guided revision
to the same fields. Within each backend, the configurations share
the index, retrieval procedure, role order, evidence processing,
and final reader. All retain basic parsing, tool-call validation,
and the Verifier on the iterative path. All six reset $M_g$ before
each query and initialize $M_t$ and $M_a$ for within-query use.
External baselines include
Vanilla RAG~\cite{lewis2020retrieval},
ReAct~\cite{yao2023react},
IRCoT~\cite{trivedi2023interleaving},
A-RAG~\cite{du2026arag},
Self-RAG~\cite{asai2024selfrag},
GraphRAG~\cite{edge2024graphrag},
DOSRAG~\cite{laitenberger2025stronger},
RAPTOR~\cite{sarthi2024raptor},
MemoRAG~\cite{qian2025memorag}, and
HippoRAG2~\cite{gutierrez2025hipporag2}.
Each baseline retains its defining indexing, retrieval, or
control procedure within a common input and final-reader interface.

All methods use Qwen2.5-14B as the final reader and Qwen2.5-7B
for auxiliary calls where required~\cite{qwen2025technical},
with greedy decoding and temperature zero.
The six controlled configurations allow at most
two retrieval rounds, 12 query-time auxiliary calls including
retries and recovery, and one reader call per query. External
baselines use method-specific control budgets. We define reader-equivalent tokens (RET) as a fixed-weight
measure of LLM input and output token usage. Reader tokens
receive unit weight, and auxiliary-model tokens receive
weight $8/14$. For a query $q$,
\begin{equation}
    \mathrm{RET}_q
    =
    T_{q,R}
    + \frac{8}{14}
      \left(T_{q,\mathrm{aux}}+T_{q,\mathrm{idx}}\right),
    \label{eq:ret}
\end{equation}
where the three terms count input and output tokens for the
final reader, query-time auxiliary calls, and query-attributable
index construction. The same weights apply to all methods.
Query-attributable indexing is charged in full.
Token counts also include retries, recovery, and all calls
from failed executions. We report mean RET per query and
retain all queries in the denominators for quality and
resource statistics. Fallback answers are scored as generated,
and missing answers receive zero.
Appendix~\ref{sec:additional_setup} provides execution
settings and token-accounting details.

\subsection{Results and Analysis}
\label{sec:results}

\paragraph{Overall performance.}
Table~\ref{tab:main_results} reports TAM Full scores of
47.8 LongBench Macro F1, 45.8 QASPER Answer F1, and
84.7 DocVQA ANLS, exceeding the strongest external baselines
by 5.3, 5.0, and 4.3 points, respectively.
On HotpotQA, 2WikiMQA, and MuSiQue, TAM Full exceeds
the strongest external baselines, Self-RAG, A-RAG, and IRCoT,
by 3.1, 5.8, and 4.4 F1 points, respectively.

\begin{table*}[!t]
\centering
\small
\setlength{\tabcolsep}{3.2pt}
\setlength{\arrayrulewidth}{0.3pt}
\renewcommand{\arraystretch}{1.15}

\resizebox{\textwidth}{!}{%
\begin{tabular}{@{}l rrr|rrr|rrr|rrr|rr|rr@{}}
\toprule[0.9pt]

\multirow{2}{*}{\textbf{Method}}
& \multicolumn{3}{c}{\textbf{HotpotQA}}
& \multicolumn{3}{c}{\textbf{2WikiMQA}}
& \multicolumn{3}{c}{\textbf{MuSiQue}}
& \multicolumn{3}{c}{\textbf{LB Avg.}}
& \multicolumn{2}{c}{\textbf{QASPER}}
& \multicolumn{2}{c}{\textbf{DocVQA}} \\

\cmidrule(lr){2-4}
\cmidrule(lr){5-7}
\cmidrule(lr){8-10}
\cmidrule(lr){11-13}
\cmidrule(lr){14-15}
\cmidrule(lr){16-17}

& EM$\uparrow$ & F1$\uparrow$ & RET$\downarrow$
& EM$\uparrow$ & F1$\uparrow$ & RET$\downarrow$
& EM$\uparrow$ & F1$\uparrow$ & RET$\downarrow$
& EM$\uparrow$ & F1$\uparrow$ & RET$\downarrow$
& F1$\uparrow$ & RET$\downarrow$
& ANLS$\uparrow$ & RET$\downarrow$ \\

\midrule
\rowcolor{tablegroup}
\multicolumn{17}{c}{\textbf{\textit{External baselines}}} \\
\midrule

Vanilla RAG
& 29.0 & 40.0 & 900
& 42.0 & 49.0 & 900
& 19.0 & 26.5 & 1,050
& 30.0 & 38.5 & 950
& 38.5 & 965
& 80.0 & 950 \\

ReAct
& 31.5 & 43.5 & 1,850
& 44.5 & 51.5 & 1,900
& 21.5 & 28.0 & 2,250
& 32.5 & 41.0 & 2,000
& 40.0 & 1,650
& 79.9 & 1,300 \\

Self-RAG
& \underline{34.5} & \underline{46.9} & 1,822
& 40.5 & 47.9 & 2,019
& 17.0 & 24.0 & 1,965
& 30.7 & 39.6 & 1,935
& 35.6 & 1,346
& 74.9 & 1,482 \\

GraphRAG
& 22.0 & 27.6 & 1,426
& 27.0 & 29.8 & 1,514
& 5.5 & 10.0 & 1,407
& 18.2 & 22.5 & 1,449
& 32.7 & 1,968
& 31.2 & 3,157 \\

DOSRAG
& 30.5 & 43.1 & 1,383
& 32.5 & 43.1 & 1,089
& 16.0 & 21.1 & 997
& 26.3 & 35.8 & 1,156
& 35.0 & 1,056
& 79.5 & 1,026 \\

RAPTOR
& 29.5 & 40.8 & 18,340
& 33.5 & 42.2 & 10,665
& 13.0 & 17.8 & 22,405
& 25.3 & 33.6 & 17,137
& 38.5 & 8,038
& 79.5 & 1,416 \\

MemoRAG
& 34.0 & 45.3 & 3,248
& 41.5 & 49.6 & 3,411
& 16.0 & 22.2 & 3,204
& 30.5 & 39.0 & 3,288
& 36.1 & 3,126
& 79.3 & 1,652 \\

HippoRAG2
& \underline{34.5} & 46.4 & 98,662
& 40.0 & 48.2 & 52,209
& 18.0 & 23.7 & 111,726
& 30.8 & 39.4 & 87,532
& 38.8 & 42,424
& 79.7 & 6,340 \\

IRCoT
& 31.5 & 43.5 & 2,450
& 45.0 & 52.0 & 2,500
& 22.0 & 29.6 & 2,850
& 32.8 & 41.7 & 2,600
& \underline{40.8} & 1,950
& 80.4 & 1,550 \\

A-RAG
& 32.5 & 44.5 & 2,950
& \underline{45.5} & 53.5 & 3,050
& \underline{22.5} & 29.5 & 3,600
& \underline{33.5} & 42.5 & 3,200
& 40.5 & 2,100
& 80.2 & 1,650 \\

\midrule
\rowcolor{tablegroup}
\multicolumn{17}{c}{\textbf{\textit{Controlled configurations}}} \\
\midrule

\rowcolor{tableblue}
\textbf{TAM Full}
& \textbf{37.5} & \textbf{50.0} & 1,768
& \textbf{51.5} & \textbf{59.3} & 1,068
& \textbf{26.5} & \textbf{34.0} & 1,326
& \textbf{38.5} & \textbf{47.8} & 1,387
& \textbf{45.8} & 1,833
& \textbf{84.7} & 1,450 \\

TAM Structured
& 32.5 & 45.6 & 1,763
& \underline{45.5} & \underline{55.4} & 1,018
& 22.0 & \underline{30.2} & 1,296
& 33.3 & \underline{43.7} & 1,359
& 40.5 & 1,750
& \underline{80.9} & 1,400 \\

TAM Free-form
& 31.0 & 42.3 & 1,869
& 43.5 & 51.0 & 1,930
& 19.5 & 27.0 & 1,857
& 31.3 & 40.1 & 1,885
& 39.5 & 1,850
& 79.2 & 1,758 \\

\midrule

Flat Full
& 32.0 & 43.0 & 1,341
& 44.5 & 52.8 & 1,007
& 21.0 & 28.5 & 1,207
& 32.5 & 41.4 & 1,185
& 40.5 & 1,550
& 80.8 & 1,250 \\

Flat Structured
& 30.0 & 42.0 & 1,205
& 43.5 & 50.8 & 1,059
& 21.0 & 27.2 & 1,157
& 31.5 & 40.0 & 1,140
& 39.8 & 1,525
& 79.9 & 1,200 \\

Flat Free-form
& 28.0 & 41.5 & 1,453
& 42.5 & 50.5 & 1,558
& 19.0 & 26.5 & 1,950
& 29.8 & 39.5 & 1,654
& 39.1 & 1,650
& 79.4 & 1,300 \\

\bottomrule[0.9pt]
\end{tabular}%
}

\caption{Results with independently processed queries. The six
controlled configurations reset $M_g$ per query. Quality scores
use a 0--100 scale. LB Avg.\ is the unweighted mean over the three
LongBench tasks. RET denotes mean reader-equivalent tokens per query.
Bold and underlined values indicate the best and second-best quality
scores, respectively.}
\label{tab:main_results}
\end{table*}

\begin{figure}[!htbp]
    \centering
    \includegraphics[width=\linewidth]{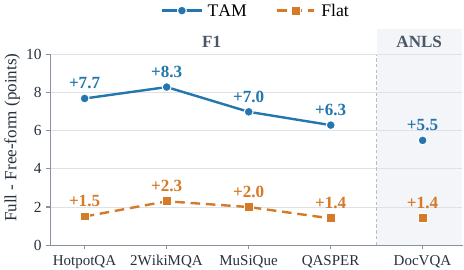}
    \caption{Full minus Free-form gains on TAM and Flat, measured
    in F1 points on LongBench and QASPER and ANLS points on DocVQA.}
    \label{fig:backend_gains}
\end{figure}

\paragraph{State representation and checked updates.}
For Free-form, Structured, and Full, LongBench Macro F1
is 40.1, 43.7, and 47.8 on TAM and 39.5, 40.0, and 41.4
on Flat (Table~\ref{tab:main_results}).
Moving from Free-form to Structured and then Full yields
3.6 and 4.1 points on TAM and 0.5 and 1.4 points on Flat,
separating the gains from explicit fields and the complete
update mechanism. TAM's advantage over Flat grows from
0.6 to 3.7 to 6.4 points across these configurations.
Figure~\ref{fig:backend_gains} plots Full--Free-form gains
by task.

On TAM, the same two changes yield 1.0 and 5.3 Answer F1
points on QASPER and 1.7 and 3.8 ANLS points on DocVQA,
respectively. Flat also improves at both steps on both
datasets. The complete update mechanism provides the
larger increment on TAM in both document tasks.

\begin{figure*}[!t]
    \centering
    \includegraphics[width=\textwidth]
        {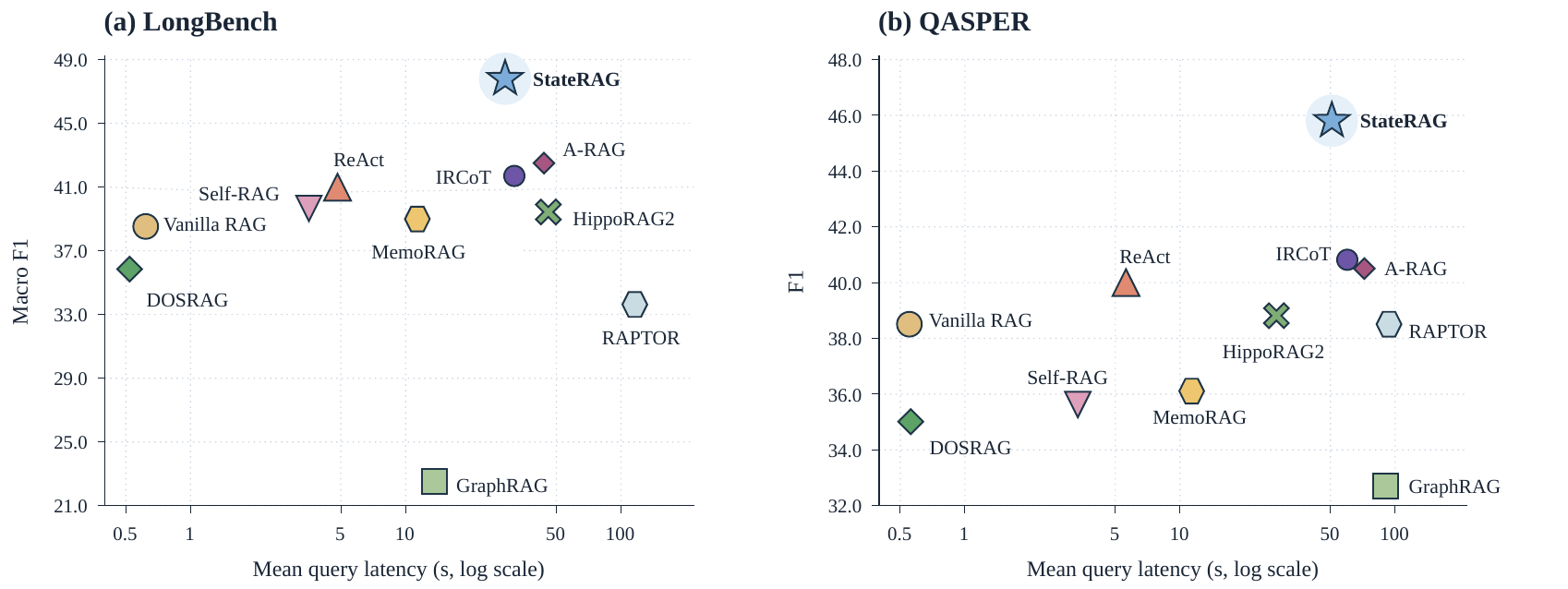}
    \caption{Answer quality versus mean query latency on LongBench
(Macro F1) and QASPER (Answer F1). Stars denote TAM Full.
Latency is shown on a logarithmic scale.}
    \label{fig:quality_latency}
\end{figure*}

\paragraph{Behavior after update rejection.}
Table~\ref{tab:rejection_analysis} compares TAM Full with TAM
Structured on the 600 LongBench queries, grouped by whether
TAM Full rejects a candidate state update.
On the 120 queries with a rejection, Full obtains 61.0 F1 and Structured
obtains 40.5, a difference of 20.5 points. Both average 44.5 F1 on the
remaining 480 queries. Weighting these group means by query count yields
the overall scores of 47.8 and 43.7. Full's average improvement is thus
concentrated in the group where update rejection occurs. Execution logs
show that 108 of the 120 queries continue after rejection, with accepted
state retained while feedback guides revision within the remaining budget.
This analysis connects the quality difference to queries where the
checked-update process intervenes.

\begin{table}[!htbp]
\centering

\setlength{\tabcolsep}{3.5pt}
\renewcommand{\arraystretch}{1.08}

\resizebox{\columnwidth}{!}{%
\begin{tabular}{@{}lrrrr@{}}
\toprule
Question group & $n$ & Full & Structured & $\Delta$ F1 \\
\midrule
Rejection triggered & 120 & 61.0 & 40.5 & $+20.5$ \\
No rejection        & 480 & 44.5 & 44.5 & $0.0$ \\
\midrule
\textbf{All questions}
& \textbf{600}
& \textbf{47.8}
& \textbf{43.7}
& $\mathbf{+4.1}$ \\
\bottomrule
\end{tabular}%
}
\caption{Mean F1 of TAM Full and TAM Structured on the same LongBench
queries, grouped by whether TAM Full rejects a candidate state update.}
\label{tab:rejection_analysis}

\end{table}

\paragraph{Component ablations.}

In \textit{w/o MARS}, we retain TAM construction and replace the iterative role workflow with cosine-similarity top-5 retrieval over TAM leaf embeddings, followed by the same final reader. This removes planning, role-based navigation, LLM-based evidence selection, verification, and feedback-driven re-retrieval. The reader model, decoding settings, and five-passage evidence cap remain unchanged. In \textit{w/o SMP}, we retain TAM and the MARS workflow, including verification and feedback, but disable the global document cache and all query-scoped and role-private memory reads and writes. Current-query evidence and feedback are passed directly through local execution variables.

Table~\ref{tab:component_ablation} shows that replacing TAM
with Flat reduces LongBench Macro F1 by 6.4 points.
Removing MARS and SMP lowers it by 9.8 and 6.6 points,
respectively. MARS removal produces the largest decrease
on both benchmarks. On QASPER, removing SMP reduces
Answer F1 by 11.2 points.
\begin{table}[!htbp]
\centering
\small
\setlength{\tabcolsep}{4pt}
\renewcommand{\arraystretch}{1.08}

\begin{tabular*}{\columnwidth}
    {@{\extracolsep{\fill}}lccc@{}}
    \toprule
    \textbf{Variant}
    & \multicolumn{2}{c}{\textbf{LongBench}}
    & \textbf{QASPER} \\
    \cmidrule(lr){2-3}
    \cmidrule(l){4-4}
    & EM$\uparrow$ & F1$\uparrow$ & F1$\uparrow$ \\
    \midrule
    w/o TAM & 32.5 & 41.4 & 40.5 \\
    w/o MARS & 29.3 & 38.0 & 34.2 \\
    w/o SMP & 32.2 & 41.2 & 34.6 \\
    \textbf{Full}
    & \textbf{38.5}
    & \textbf{47.8}
    & \textbf{45.8} \\
    \bottomrule
\end{tabular*}
\caption{Component ablations with $M_g$ reset per query.
The TAM ablation uses the Flat backend. The SMP ablation
disables storage and reuse of intermediate retrieval
results within each query.}
\label{tab:component_ablation}
\end{table}

\paragraph{Global memory reuse.}

Table~\ref{tab:global_memory} evaluates the additional effect of retaining
$M_g$ across a fixed LongBench query sequence. Both settings use the
same queries, order, models, and budgets, starting with empty global
memory. Relative to per-query reset, retention raises Macro F1 from
47.8 to 48.4 and reduces mean RET from 1,387 to 1,330, a decrease of
57 RET or 4.1\%.
F1 increases by 0.2, 1.2, and 0.4 points on HotpotQA, 2WikiMQA, and
MuSiQue, respectively. Cross-query artifact reuse therefore provides
an additional gain in this sequential workload.

\begin{table}[!htbp]
    \centering
    \small
    \setlength{\tabcolsep}{3pt}
    \renewcommand{\arraystretch}{1.12}

    \begin{tabular*}{\columnwidth}
        {@{\extracolsep{\fill}}lccccc@{}}
        \toprule
        \textbf{Setting}
        & \textbf{Hotpot}
        & \textbf{2Wiki}
        & \textbf{MuSiQue}
        & \textbf{Macro}
        & \textbf{RET} \\
        \midrule
        $M_g$-Reset
        & 50.0 & 59.3 & 34.0 & 47.8 & 1,387 \\
        $M_g$-Retained
        & 50.2 & 60.5 & 34.4 & 48.4 & 1,330 \\
        \midrule
        \textbf{Difference}
        & \textbf{+0.2}
        & \textbf{+1.2}
        & \textbf{+0.4}
        & \textbf{+0.6}
        & $\mathbf{-57}$ \\
        \bottomrule
    \end{tabular*}

    \caption{Global memory reuse in TAM Full. $M_g$ is reset per
    query or retained across a fixed LongBench query sequence.
    Task scores are F1, and Macro is their mean.}
    \label{tab:global_memory}
\end{table}

\paragraph{Resource usage.}
Table~\ref{tab:main_results} reports RET alongside answer quality. On LongBench, TAM Full uses 1,387 RET per query, compared with 1,359 for Structured and 1,885 for Free-form. Its 4.1-point gain over Structured accompanies an increase of 28 RET per query, including recovery calls. Figure~\ref{fig:quality_latency} complements these token measurements with mean query latency on LongBench and QASPER.

\section{Conclusion}
\label{sec:conclusion}

We presented StateRAG, which manages multi-step retrieval through
explicit state fields. Roles propose updates to this state, and a controller checks them
before acceptance. When an update is rejected, the controller retains the
accepted state and uses feedback to guide revision within the query budget.
Controlled comparisons with independently processed queries support the value
of explicit fields and show further improvements from the full update
mechanism on both TAM and Flat. Comparisons with iterative retrieval baselines
and document QA experiments further support the performance of the complete
system. A separate sequential workload analysis shows additional benefits
from cross-query artifact reuse.

\section*{Limitations}

Our evaluation uses one reader--auxiliary model pair with fixed retrieval and call budgets. The controlled comparisons assess update checking, accepted-state retention, and feedback-guided revision as a combined mechanism. DocVQA results concern retrieval and answering over a shared image-to-text representation.
Global-memory reuse is evaluated on a fixed query sequence, and its benefits may vary with query relatedness and the availability of reusable artifacts.


\FloatBarrier
\bibliography{custom}

\clearpage
\appendix

\section{Additional Experimental Details}
\label{sec:additional_setup}

\subsection{Evaluation Data and Shared Inputs}

The LongBench evaluation contains 200 queries each from
HotpotQA, 2WikiMQA, and MuSiQue, for 600 queries in total.
We also evaluate the 1,451-query QASPER test set and
2,000 fixed DocVQA validation queries.
LongBench reports EM and F1, with an unweighted mean
across the three tasks for each metric.
QASPER reports Answer F1, and DocVQA reports ANLS.
Quality scores use a 0--100 scale.

All methods receive the same questions and source materials.
For DocVQA, Qwen-VL-Plus converts document images into
textual regions with page and region identifiers.
The cached conversion outputs are shared across methods.
The final reader receives the question and selected
textual evidence.

\subsection{Execution Settings}
\label{app:execution_settings}

Table~\ref{tab:appendix_execution_settings} lists the
models, environment, and execution limits.
Within each backend, Free-form, Structured, and Full
share the index, retrieval procedure, role order,
evidence processing, final reader, and query budgets.
The two backends use the same segmentation protocol
and \textsc{Bypass} rule. For each configuration, the reported results come from a
single evaluation run on the fixed query set.
Task scores aggregate query-level results from that run,
and LongBench macro scores are unweighted means across
the three tasks.

\begin{table*}[t]
    \centering
    \small
    \setlength{\tabcolsep}{5pt}
    \renewcommand{\arraystretch}{1.12}
    \begin{tabularx}{\textwidth}{
        @{}p{0.29\textwidth}
        >{\raggedright\arraybackslash}X@{}
    }
        \toprule
        Item & Setting \\
        \midrule

        Final reader
        & Qwen2.5-14B \\

        Auxiliary model
        & Qwen2.5-7B \\

        DocVQA front end
        & Qwen-VL-Plus with shared cached text outputs \\

        Decoding
        & Greedy decoding with temperature 0 \\

        Hardware
        & Local server with NVIDIA RTX 3090 GPUs \\

        Runtime
        & Python 3.10, PyTorch, and
          Hugging Face Transformers \\

        Retrieval rounds
        & At most two per query \\

        Query-time auxiliary calls
        & At most 12 per query, including path scoring,
          role calls, ordinary retries, reassessment,
          and recovery \\

        Full recovery
        & At most two episodes, each using at most
          two auxiliary calls within the 12-call limit \\

        Final-reader calls
        & At most one per query, separate from
          the auxiliary-call limit \\

        Output limits
        & 512 tokens per auxiliary call and
          1,024 tokens per reader call \\

        Historical context
        & At most 24,000 characters,
          retaining the tail after truncation \\

        Final evidence
        & At most five passages within a
          2,048-token evidence-input budget \\

        Main controlled comparisons
        & Reset $M_g$ before each query;
          initialize $M_t$ and $M_a$ for within-query use \\

        \bottomrule
    \end{tabularx}
    \caption{Models and execution settings.
    Retrieval, call, recovery, context, and passage-count
    limits apply to the six controlled configurations.
    External baselines use method-specific control budgets.}
    \label{tab:appendix_execution_settings}
\end{table*}

\paragraph{Budget accounting.}
The 12-call limit covers query-time auxiliary-model
requests, including calls made during unsuccessful
execution attempts.
Recovery calls draw from this same allowance.
The two recovery episodes are a query-level limit,
with at most two auxiliary calls in each episode.
Recovery does not reset the retrieval-round or
auxiliary-call counters.

The final reader has a separate allowance of one call
across \textsc{Bypass}, \textsc{Release}, and
\textsc{Fallback}.
Index-construction calls are outside the query-time
auxiliary-call limit and are included in resource
accounting.

\paragraph{Context and evidence limits.}
Historical context is truncated to 24,000 characters
by retaining its tail.
The current question, fixed role instructions, and
required state inputs are supplied separately from
this truncated history.

The evidence token count uses the final reader's
tokenizer and the serialized evidence input,
including separators.
The 2,048-token evidence budget is distinct from
the reader's 1,024-token output limit.
Evidence preparation applies both the passage-count
and token limits before the snapshot is assessed.

\paragraph{Bypass conditions.}
Both backends require nonempty initial evidence,
at most five evidence units in the complete accessible
source collection, and initial evidence that fits
the reader's evidence token budget.
The source-unit count is computed over the complete
accessible collection.
Units are TAM Evidence leaves or Flat passages
under the common segmentation protocol.

The token count is computed over the initial evidence
serialized in the reader's evidence format, including
separators, before any top-$k$ or token-budget truncation.
Eligible queries pass this initial evidence directly
to the reader in retrieval order.

\paragraph{Termination and scoring.}
On the iterative path, \textsc{Release} sends the assessed
evidence snapshot unchanged to the reader.
When further retrieval or revision cannot proceed,
\textsc{Fallback} selects available accepted evidence
within the reader's limits.
Execution ends without a reader call if the required
inputs cannot be formed.

Fallback answers are scored as generated, and missing
answers receive zero.
All evaluated queries remain in the denominators
for quality and resource statistics.

\subsection{Baseline Alignment and Component Ablations}

External baselines retain their defining indexing,
retrieval, reflection, or control procedures.
Their selected evidence is converted to the common
final-reader format.
The reader receives the question and evidence,
while intermediate reasoning and control trajectories
remain internal to the baseline.
Table~\ref{tab:appendix_baseline_alignment} summarizes
this alignment.

\begin{table*}[t]
    \centering
    \small
    \setlength{\tabcolsep}{4pt}
    \renewcommand{\arraystretch}{1.12}
    \begin{tabularx}{\textwidth}{
        @{}p{0.16\textwidth}
        >{\raggedright\arraybackslash}X
        >{\raggedright\arraybackslash}X@{}
    }
        \toprule
        Method
        & Preserved procedure
        & Evidence passed to the reader \\
        \midrule

        Vanilla RAG
        & One similarity-based passage retrieval pass
        & Top-ranked retrieved passages \\

        ReAct
        & Interleaved reasoning, retrieval actions,
          and observations
        & Evidence collected during the interaction \\

        IRCoT
        & Retrieval interleaved with intermediate
          chain-of-thought reasoning
        & Evidence collected across retrieval steps \\

        A-RAG
        & Agent-directed evidence acquisition through
          hierarchical retrieval interfaces
        & Evidence selected by the retrieval agent \\

        Self-RAG
        & Reflection decisions controlling retrieval
          and passage assessment
        & Evidence retained by the reflection procedure \\

        GraphRAG
        & Entity-graph construction, community detection,
          and community summaries
        & Evidence retrieved from graph and
          community representations \\

        DOSRAG
        & Passage retrieval followed by restoration
          to the original document order
        & Selected passages in source order \\

        RAPTOR
        & Recursive embedding, clustering, and
          summarization over multiple abstraction levels
        & Retrieved nodes from the abstraction hierarchy \\

        MemoRAG
        & Corpus-memory construction and
          memory-generated clues for retrieval
        & Evidence selected through clue-guided retrieval \\

        HippoRAG2
        & Passage and relation graphs, recognition
          filtering, and Personalized PageRank
        & Evidence selected by the graph-retrieval procedure \\

        \bottomrule
    \end{tabularx}
    \caption{Baseline procedures and alignment with
    the common final-reader interface.}
    \label{tab:appendix_baseline_alignment}
\end{table*}

External methods use their method-specific control
budgets. The two-round and 12-auxiliary-call limits
define the six controlled configurations.
The selected baseline evidence is subject to the
shared reader evidence-token budget.

\paragraph{Removing TAM.}
The \textit{w/o TAM} variant is Flat Full.
It replaces graph traversal with passage retrieval
while retaining the Full state-update workflow.

\paragraph{Removing MARS.}
The \textit{w/o MARS} variant retains per-query TAM
construction and replaces iterative role execution
with cosine-similarity top-5 retrieval over TAM
leaf embeddings, followed by the same final reader.
It removes planning, role-based navigation,
LLM-based evidence selection, verification,
and feedback-driven re-retrieval.
Reader settings and the five-passage evidence cap
remain unchanged.
This variant performs one retrieval pass, so the
MARS iteration limit does not apply.

\paragraph{Removing SMP.}
The \textit{w/o SMP} variant retains TAM and the MARS
workflow, including evidence verification and feedback.
It disables the document cache and all query-scoped
and role-private memory reads and writes.
Current-query evidence and feedback remain available
through local execution variables.
The component comparisons reset $M_g$ per query.

\subsection{RET and Latency Accounting}

For each query, we aggregate input and output tokens
from three disjoint categories: final-reader calls,
query-time auxiliary calls, and query-attributable
index-construction calls.
Each call contributes once to its corresponding
category in Equation~\ref{eq:ret}.
Qwen2.5-14B calls receive weight 1, while Qwen2.5-7B
auxiliary and indexing calls receive weight $8/14$.
These fixed weights apply to all methods.

Query-attributable indexing is included in full.
Token counts also include ordinary retries, recovery,
and all calls from failed executions.
We calculate RET for each query and average over
all evaluated queries, including those without
a generated answer.
RET covers LLM token usage.
Embedding computation and non-LLM graph processing
fall outside this count.

Latency is measured from query submission to final
answer return or recorded termination.
It includes query-attributable indexing and online
execution.
The arithmetic mean retains elapsed time from
incomplete executions and uses all evaluated queries
as its denominator.
Figure~\ref{fig:quality_latency} reports measured
query latency separately from RET.

\subsection{Sequential Memory Reuse}
\label{app:sequential_memory}

The sequential analysis compares TAM Full with
per-query $M_g$ reset against TAM Full with $M_g$
retained across a fixed LongBench query sequence.
Both settings use the same queries, order, models,
and execution budgets and start with empty global
memory.
The retention setting carries eligible stored results
between queries.
Query-scoped memory $M_t$ and role-private memory $M_a$
are reinitialized for every query in both settings.

Reusable objects are described in
Appendix~\ref{app:smp_details}.
Their source references must remain within the
materials accessible to the current query.
Reference answers, evaluation labels, and
final-reader outputs are excluded from global memory.

Resource accounting includes the calls that initially
produce stored results and any calls executed after
a cache hit.
A generation call skipped through reuse contributes
no tokens to that query.
The same fixed RET weights and complete-query
denominator are used in both memory settings.

\section{Retrieval State and Execution Example}
\label{app:state_example}

We illustrate the retrieval workflow with the question,
``Who was born first, Marie Curie or Albert Einstein?''
After obtaining evidence for Curie's birth year, the system
records the missing Einstein birth year in its assessment.
This gap guides the next plan and retrieval step. The new
passage is combined with the retained evidence to form a
snapshot for reassessment.
Table~\ref{tab:appendix_state_example} shows the accepted
state after each assessment.

\paragraph{Example setting.}
The example follows the iterative retrieval path.
The accessible source collection contains more than five
evidence units, so the source-unit condition for
\textsc{Bypass} is not met.
Only the relevant nodes and passages are shown.
Node, passage, and source identifiers are local labels
used for this illustration, and passage text is abbreviated
to the facts needed for the comparison.

\begin{table*}[t]
    \centering
    \small
    \setlength{\tabcolsep}{5pt}
    \renewcommand{\arraystretch}{1.15}
    \begin{tabularx}{\textwidth}{
        @{}l
        >{\raggedright\arraybackslash}X
        >{\raggedright\arraybackslash}X@{}
    }
        \toprule
        Field
        & After the first assessment
        & After further retrieval and assessment \\
        \midrule

        Plan $p$
        & Compare Marie Curie's and Albert Einstein's
          birth years.
        & Find Albert Einstein's birth year. \\

        Retrieval trace $\pi$
        & Root \texttt{root}
          $\rightarrow$ Topic \texttt{scientists}
          $\rightarrow$ Relation \texttt{curie\_bio}
          $\rightarrow$ Evidence \texttt{region\_c}.
        & Root \texttt{root}
          $\rightarrow$ Topic \texttt{scientists}
          $\rightarrow$ Relation \texttt{einstein\_bio}
          $\rightarrow$ Evidence \texttt{region\_e}. \\

        Evidence $E$
        & \texttt{passage\_c}, from \texttt{doc\_c}:
          Marie Curie was born in 1867.
        & Retain \texttt{passage\_c}.
          Add \texttt{passage\_e}, from \texttt{doc\_e}:
          Albert Einstein was born in 1879. \\

        Assessment $z$
        & \textsc{Fail}: Einstein's birth year
          is missing.
        & \textsc{Pass}: both birth years
          are available for comparison. \\

        Artifact refs $U$
        & \emph{Omitted}.
        & \emph{Omitted}. \\

        \bottomrule
    \end{tabularx}
    \caption{Illustrative state evolution during retrieval.
    Each column shows an accepted state after assessment.
    Stored-result references are omitted from the display.}
    \label{tab:appendix_state_example}
\end{table*}

\paragraph{Evidence records.}
An Evidence node in TAM locates a source region.
The Retriever selects passages from that region and
preserves their source information in $E$.
For example, an abbreviated record for the first passage is:

\begin{quote}
\small
\raggedright
\texttt{passage\_id}: \texttt{passage\_c}\\
\texttt{source\_id}: \texttt{doc\_c}\\
\texttt{region\_id}: \texttt{region\_c}\\
\texttt{text}: Marie Curie was born in 1867.
\end{quote}

The current trace $\pi$ changes when a new retrieval choice
is accepted. Evidence retained from an earlier round keeps
its original source information. In the second column,
\texttt{passage\_c} therefore remains in $E$ with its original
provenance, while \texttt{passage\_e} is obtained through the
newly accepted path. References in $U$ identify stored
intermediate results and are omitted here to keep the
display focused on retrieval and assessment.

\paragraph{From assessment to the next action.}
The first \textsc{Fail} assessment is accepted into $z$
and identifies the information still needed.
When the remaining budget permits another retrieval round,
the Planner uses this assessment to target Einstein's
birth year. The Navigator proposes a path to the relevant
region, and the Retriever selects the additional passage.
Each role's proposal passes through the update-check
procedure before changing the accepted state.

The new passage is combined with retained evidence,
deduplicated, ranked, and truncated to the evidence budget
before the resulting snapshot is proposed for $E$.
Once this update is accepted, the Verifier assesses that
snapshot and proposes a new assessment.
After the \textsc{Pass} assessment is accepted, the
controller selects \textsc{Release}.
The reader receives the question and the same evidence
snapshot, compares 1867 with 1879, and answers
``Marie Curie.''

The two columns in Table~\ref{tab:appendix_state_example}
are separated by several field updates.
The state index $t$ counts update attempts, including
rejected attempts, rather than retrieval rounds.
A rejected update leaves the accepted state unchanged.

\paragraph{Flat retrieval records.}
With the Flat backend, $\pi$ stores the retrieval request
and returned ranking. Its fields are
\texttt{backend}, \texttt{round\_id},
\texttt{index\_id}, \texttt{retrieval\_query},
\texttt{top\_k}, and \texttt{hits}.
Each hit contains a \texttt{passage\_id},
\texttt{rank}, and \texttt{score}.
The Navigator assembles this record from the request
and backend results. After the record is accepted,
the Retriever obtains passage text by ID and forms
the candidate evidence snapshot.
Selected passages are stored in $E$, while earlier
retrieval records remain in the execution log.

\section{Update Checks and Revision}
\label{app:update_checks}

\subsection{Shared Execution Requirements}

Free-form, Structured, and Full retain basic parsing,
tool-call validation, identifier and access checks,
and the Verifier on the iterative path.
They also share the evidence-processing and reader-admission
rules. Evidence is prepared before assessment, and
\textsc{Release} passes the assessed snapshot unchanged
to the reader.
Table~\ref{tab:appendix_carrier_controls} summarizes
the differences among the three configurations.

\begin{table*}[t]
    \centering
    \small
    \setlength{\tabcolsep}{4pt}
    \renewcommand{\arraystretch}{1.12}
    \begin{tabularx}{\textwidth}{
        @{}l
        >{\raggedright\arraybackslash}X
        >{\raggedright\arraybackslash}X
        >{\raggedright\arraybackslash}X@{}
    }
        \toprule
        Item & Free-form & Structured & Full \\
        \midrule

        Intermediate information
        & Append-only text trajectory with role labels
        & Explicit state fields
        & Same explicit state fields \\

        Basic execution checks
        & Parsing, tool-call validation, identifier
          validity, and access control
        & Same
        & Same \\

        Evidence assessment
        & Verifier on the iterative path
        & Same
        & Same \\

        Evidence lifecycle
        & Prepare the snapshot before assessment;
          preserve it on \textsc{Release}
        & Same
        & Same \\

        Additional update checks
        & Disabled
        & Disabled
        & Check associations between proposed updates
          and accepted state \\

        Extra-check rejection
        & Not triggered
        & Not triggered
        & Retain accepted state and revise recoverable
          proposals using check feedback \\

        Insufficient evidence
        & Continue retrieval when budget permits
        & Same
        & Same \\

        \bottomrule
    \end{tabularx}
    \caption{State presentation and update handling.
    The same comparison applies within TAM and Flat.}
    \label{tab:appendix_carrier_controls}
\end{table*}

Fixed role scheduling assigns updates to their designated
sources. The Planner, Navigator, Retriever, and Verifier
propose changes to $p$, $\pi$, $E$, and $z$, respectively.
SMP supplies the stored-result references in $U$.
This scheduling constraint applies throughout the workflow.

\subsection{Additional Checks in Full}

Full checks the associations declared by a candidate update
before accepting it. An identifier may resolve to a valid
object while referring to the wrong plan version, retrieval
record, or stored result.
These associations determine whether the proposal is
consistent with the state available at that step.
Table~\ref{tab:appendix_update_checks} distinguishes
the shared checks from these additional requirements.

\begin{table*}[t]
    \centering
    \small
    \setlength{\tabcolsep}{4pt}
    \renewcommand{\arraystretch}{1.12}
    \begin{tabularx}{\textwidth}{
        @{}p{0.19\textwidth}
        >{\raggedright\arraybackslash}X
        p{0.15\textwidth}@{}
    }
        \toprule
        Check & Requirement & Applied to \\
        \midrule

        Basic parsing
        & Required response items and tool arguments
          are parseable and usable.
        & All three \\

        Identifier validity
        & Referenced nodes, passages, and stored
          objects exist and can be resolved.
        & All three \\

        Access control
        & Access is permitted for the requesting
          role and memory scope.
        & All three \\

        Assessment format
        & The Verifier returns a valid
          \textsc{Pass} or \textsc{Fail} verdict
          with a justification.
        & All three \\

        Plan association
        & A proposal's referenced plan version
          matches the accepted plan used for
          the current retrieval step.
        & Full \\

        Retrieval association
        & Newly added evidence agrees with its
          declared retrieval record and source.
          It belongs to the selected region or
          returned candidate set.
        & Full \\

        Stored-object association
        & References declared in a proposal match
          the stored objects and source associations
          they identify.
        & Full \\

        \bottomrule
    \end{tabularx}
    \caption{Check requirements and their scope.
    Full applies the additional association checks
    before accepting candidate state updates.}
    \label{tab:appendix_update_checks}
\end{table*}

The current-retrieval requirement applies to newly added
evidence. Passages retained from earlier rounds preserve
their original source and retrieval associations.
They can remain in $E$ when the current trace changes.

\subsection{Rejection Feedback and Revision}

The update checks are deterministic procedures.
When a check fails, the controller produces structured
feedback containing the target field, error type,
observed association, and required association.
The rejected proposal does not overwrite the accepted state.

For a recoverable proposal, the controller returns this
feedback to the producing role together with its original
task, the accepted state, the rejected proposal,
available candidates and references, and the remaining
budget. The role revises its proposal, which then undergoes
the same checks.
The revision input is given in
Appendix~\ref{app:role_prompts}.

\paragraph{Illustrative rejection.}
Suppose the accepted retrieval record is
\texttt{trace\_b}.
A proposed evidence update adds
\texttt{passage\_a}, which is readable and accessible,
but declares that it was obtained through
\texttt{trace\_b}.
The retrieval record instead associates this passage
with \texttt{trace\_a}.
The identifiers pass the shared validity checks,
while Full rejects the inconsistent association.
Using local example labels, the sequence is:

\begin{quote}
\small
\raggedright
\textbf{Accepted retrieval record:}
\texttt{trace\_b}.\\
\textbf{Rejected proposal:}
add \texttt{passage\_a} with source record
\texttt{trace\_b}.\\
\textbf{Check feedback:}
the passage was obtained through \texttt{trace\_a};
the proposed association requires a candidate
returned by \texttt{trace\_b}.\\
\textbf{Revision input:}
the accepted state, rejected proposal, feedback,
and candidates returned by \texttt{trace\_b}.\\
\textbf{Revised proposal:}
select \texttt{passage\_b} from that candidate set
and retain its source association.
\end{quote}

In this illustration, revision changes the selected
passage. The revised evidence is prepared under the
common evidence budget, submitted as an update,
and assessed after acceptance.

\subsection{Recovery Budgets and Subsequent Actions}

A proposal is recoverable when the reported inconsistency
can be corrected through evidence selection, retrieval,
or a revised update within the remaining budget.
Full permits at most two recovery episodes per query,
with at most two auxiliary calls per episode.
These calls are included in the shared limit of
12 query-time auxiliary calls.
Ordinary retries and further evidence assessments
also count toward this limit.
The retrieval-round limit remains two, and the final
reader has a separate limit of one call per query.

An accepted \textsc{Fail} assessment records insufficient
evidence and guides another retrieval round when the
budget permits. Update rejection instead requests
correction of a candidate proposal while preserving
accepted state.
Whenever accepted evidence changes, the Verifier
assesses the new snapshot before the controller
uses its verdict to select the next action.

If the remaining budget cannot support recovery,
or no admissible proposal can be formed, the controller
applies the common terminal policy.
\textsc{Fallback} uses available accepted evidence
within the reader's limits.
Execution ends without a reader call when the required
inputs cannot be formed.

\subsection{Complete Execution Procedure}

Algorithm~\ref{alg:staterag_execution} gives the Full workflow
for either retrieval backend. The budget record $b$ tracks
retrieval rounds, auxiliary calls, and recovery episodes.
All query-time auxiliary requests, including scoring, retries,
and failed requests, use the same call counter. A recovery
request also uses its episode's call counter $u$.
Budgeted proposal and revision procedures return $\bot$ when
they cannot produce a proposal within the remaining allowance.
Index construction is accounted for separately.

\begin{algorithm}[!t]
\caption{StateRAG execution with checked updates}
\label{alg:staterag_execution}
\footnotesize
\begin{algorithmic}[1]
\Require Query $q$, retrieval backend $\mathcal{B}$,
         memory pool $M$ under the selected reset policy
\Ensure Answer $y$ or incomplete result $\bot$

\State $b \gets \Call{Budget}{2,12,2}$
\State $(s,E_0) \gets \Call{Initialize}{q,\mathcal{B},M,b}$

\If{$\Call{BypassEligible}{E_0,\mathcal{B}}$}
    \State \Return $\Call{Reader}{q,E_0}$
        \Comment{\textsc{Bypass}}
\EndIf

\While{$\Call{CanStartRound}{b}$}
    \State $b.\mathrm{rounds} \gets b.\mathrm{rounds}+1$
    \State $\mathit{complete} \gets \mathrm{true}$

    \For{$R$ in $(\mathrm{Planner},\mathrm{Navigator},
                   \mathrm{Retriever},\mathrm{Verifier})$}
        \State $(s,\mathit{ok}) \gets
            \Call{CheckedUpdate}{R,q,s,\mathcal{B},M,b}$
        \If{$\neg\mathit{ok}$}
            \State $\mathit{complete} \gets \mathrm{false}$
            \State \textbf{break}
        \EndIf
    \EndFor

    \If{$\neg\mathit{complete}$}
        \State \textbf{break}
    \EndIf

    \If{$s.z.\mathrm{verdict}=\textsc{Pass}$}
        \State \Return $\Call{Reader}{q,s.E}$
            \Comment{\textsc{Release}: use the assessed snapshot}
    \EndIf

    \State Use the accepted \textsc{Fail} justification
        in the next plan \Comment{\textsc{Revise}}
\EndWhile

\State $E^\star \gets \Call{FallbackEvidence}{s}$
\If{$\neg\Call{ReaderInputsAvailable}{q,E^\star}$}
    \State \Return $\bot$
\EndIf
\State \Return $\Call{Reader}{q,E^\star}$
    \Comment{\textsc{Fallback}}

\Statex
\Function{CheckedUpdate}{$R,q,s,\mathcal{B},M,b$}
    \State $\Delta \gets
        \Call{BudgetedPropose}{R,q,s,\mathcal{B},M,b}$
    \State $u \gets \bot$
        \Comment{No recovery episode has started for this proposal}

    \While{$\Delta\neq\bot$}
        \If{$R=\mathrm{Retriever}$}
            \State $\Delta \gets
                \Call{PrepareEvidenceUpdate}{\Delta,s.E,b}$
        \EndIf

        \State $(\mathit{valid},f) \gets \Call{Check}{s,\Delta}$

        \If{$\mathit{valid}$}
            \State $s \gets \Call{Apply}{s,\Delta}$
        
            \State \Return $(s,\mathrm{true})$
        \EndIf

        \If{$\neg\Call{Recoverable}{\Delta,f,b}$}
            \State \Return $(s,\mathrm{false})$
        \EndIf

        \If{$u=\bot$}
            \If{$b.\mathrm{episodes}=2$}
                \State \Return $(s,\mathrm{false})$
            \EndIf
            \State $b.\mathrm{episodes}
                \gets b.\mathrm{episodes}+1$, $u\gets0$
        \EndIf

        \State $\Delta \gets
            \Call{BudgetedRevise}{R,q,s,\Delta,f,\mathcal{B},M,b,u}$
    \EndWhile

    \State \Return $(s,\mathrm{false})$
\EndFunction
\end{algorithmic}
\end{algorithm}

The counters $b$ and $u$ are shared with the called procedures.
Before dispatch, each auxiliary request requires
$b.\mathrm{calls}<12$ and increments that counter.
A recovery request additionally requires $u<2$ and increments $u$.
Any retrieval round started during recovery consumes the same
two-round allowance.
\textsc{BudgetedRevise} uses the original role inputs, accepted
state, rejected proposal, check feedback, available candidates,
and remaining budget specified in Appendix~\ref{app:role_prompts}.
Rejected proposals leave $s$ unchanged.

\textsc{PrepareEvidenceUpdate} combines new passages with retained
evidence, deduplicates, ranks, and applies the five-passage and
2,048-token limits. It is applied to both initial and revised
evidence proposals. Retained passages keep their original source
and retrieval associations. The Verifier evaluates the accepted
snapshot, and \textsc{Release} is reached only after its assessment
update is accepted. 

\textsc{BypassEligible} uses the complete accessible source-unit
count and the untruncated initial evidence, as specified in
Appendix~\ref{app:execution_settings}.
\textsc{FallbackEvidence} selects available accepted evidence
within the same reader limits. Each answer branch returns
immediately, so the reader is called at most once, separately
from the auxiliary-call allowance.

\section{Role Prompts and Interfaces}
\label{app:role_prompts}

The retrieval roles share their task instructions across
Free-form, Structured, and Full.
Free-form presents intermediate information as a chronological
trajectory with role labels. Structured and Full present
the corresponding information through explicit state fields.
The templates below show the task instructions and output
wrappers used by Structured and Full.
Free-form appends role responses to the trajectory while
retaining the parsing required for tool arguments, selected
evidence identifiers, and Verifier verdicts.

Table~\ref{tab:appendix_prompt_interfaces} summarizes
the interfaces. Braced placeholders are filled with
the context available to the component at execution time.

\begin{table*}[t]
    \centering
    \small
    \setlength{\tabcolsep}{4pt}
    \renewcommand{\arraystretch}{1.12}
    \begin{tabularx}{\textwidth}{
        @{}p{0.19\textwidth}
        >{\raggedright\arraybackslash}X
        >{\raggedright\arraybackslash}X@{}
    }
        \toprule
        Component & Input & Output \\
        \midrule

        Planner
        & Question, current retrieval context,
          available references, and assessment feedback
        & \texttt{OBJECTIVE} and \texttt{TARGETS} \\

        TAM Navigator and scorer
        & Question, current plan, and admissible
          candidate traces
        & Scorer: \texttt{TRACE\_ID} and \texttt{SCORE}.
          Navigator: a complete retrieval record $\pi$ \\

        Retriever
        & Question, plan, accepted retrieval choice,
          and candidate evidence
        & Ranked \texttt{SELECTED} passage identifiers \\

        Verifier
        & Question and accepted evidence snapshot
        & \texttt{VERDICT} and \texttt{JUSTIFICATION} \\

        Final reader
        & Question and selected evidence
        & A concise supported answer \\

        TAM abstraction generator
        & Node type and child-node text
        & \texttt{ABSTRACTION} \\

        DocVQA front end
        & Document image
        & \texttt{PAGE}, \texttt{REGION\_ID},
          and \texttt{TEXT} \\

        \bottomrule
    \end{tabularx}
    \caption{Inputs and outputs for retrieval roles,
    answer generation, and preprocessing.}
    \label{tab:appendix_prompt_interfaces}
\end{table*}

\subsection{Retrieval and Answering Prompts}

\paragraph{Planner.}
The Planner identifies the information needed for the next
retrieval step. Following an accepted \textsc{Fail} assessment,
the Verifier's justification supplies the remaining evidence
gap. Update-rejection feedback is provided through the
separate revision block below.

\begin{quote}
\small
\raggedright
You are the Planner in a retrieval system.
Define the current retrieval objective and the information
targets needed to answer the query.
Use the current retrieval context and any available
stored results.

When assessment feedback is present, update the plan
to address the information identified as missing
by the Verifier.
Do not select evidence, assess evidence sufficiency,
answer the query, or choose a controller action.

Query: \{query\}\\
Current retrieval context: \{role context\}\\
Available stored-result references: \{references\}\\
Assessment feedback:
\{Verifier justification or none\}

Output:\\
\texttt{OBJECTIVE:}
\{current retrieval objective\}\\
\texttt{TARGETS:}
\{ordered retrieval targets\}
\end{quote}

\paragraph{TAM Navigator and trace scorer.}
The backend constructs admissible candidate traces before
they are presented to the scorer.
The scorer evaluates these existing candidates using
the question, current plan, and node abstractions.

\begin{quote}
\small
\raggedright
You score candidate traversal traces in a typed retrieval
hierarchy.
Assess how well the terminal abstraction of each trace
supports the retrieval objective expressed by the query
and current plan.

Assign one score between 0 and 1 to every supplied candidate.
A higher score indicates that continuing along the trace
is more likely to reach evidence needed for the query.
Use only the supplied traces.
Do not create node identifiers, alter the node order,
retrieve evidence, or answer the query.

Query: \{query\}\\
Current plan: \{plan\}\\
Candidate traces:
\{trace identifiers and ordered node abstractions\}

Output for each candidate:\\
\texttt{TRACE\_ID:} \{identifier\},
\texttt{SCORE:} \{score\}
\end{quote}

The scorer's output provides candidate scores.
The Navigator resolves each \texttt{TRACE\_ID} to its
existing candidate record and selects a completed admissible
path under the navigation limits.
It then assembles the proposal for $\pi$ from the selected
node sequence and retrieval metadata.
The node sequence comes from the candidate record.

For Flat retrieval, the Navigator submits a passage-retrieval
query and collects ranked hits.
It assembles $\pi$ from the request, returned passage
identifiers, ranks, scores, and backend metadata, as described
in Appendix~\ref{app:state_example}.

\paragraph{Retriever.}
The Retriever receives candidates associated with the
accepted retrieval choice and selects passages for
the current information need.

\begin{quote}
\small
\raggedright
You are the Retriever in a retrieval system.
Rank the supplied candidate evidence by its relevance
to the query and current retrieval plan.
Select only evidence that contributes information needed
to satisfy the current objective.

Use only the supplied candidate evidence.
Preserve the evidence identifiers and source provenance.
Do not introduce external information, assess final
evidence sufficiency, or answer the query.
Select no more than the stated evidence cap.

Query: \{query\}\\
Current plan: \{plan\}\\
Accepted retrieval choice:
\{path or retrieval record\}\\
Candidate evidence:
\{candidate passages with provenance\}\\
Evidence cap: \{configured cap\}

Output:\\
\texttt{SELECTED:}
\{evidence identifiers in ranked order\}
\end{quote}

Selected identifiers are resolved to passage text and
source records. New passages are combined with retained
evidence, deduplicated, ranked, and truncated before
the resulting snapshot is proposed as an update to $E$.

\paragraph{Verifier.}
The Verifier assesses the accepted evidence snapshot.
Its verdict and justification form the proposed update
to $z$.

\begin{quote}
\small
\raggedright
You are the Verifier in a retrieval system.
Determine whether the supplied evidence is sufficient
to answer the query.

Return Pass only when the evidence contains the information
needed to produce a supported answer.
Return Fail when a required fact, relation, comparison,
or supporting step is missing.
For Fail, identify the missing information that the next
retrieval cycle should seek.

Judge evidence sufficiency only.
Do not answer the query and do not select Release,
Revise, Fallback, or Bypass.

Query: \{query\}\\
Accepted evidence: \{evidence with provenance\}

Output:\\
\texttt{VERDICT:} \{Pass or Fail\}\\
\texttt{JUSTIFICATION:}
\{brief evidence assessment or missing information\}
\end{quote}

After the assessment update is accepted, the controller
uses its verdict and the remaining budget to select
the next action.

\paragraph{Final reader.}
All methods use the same final-reader instruction.
The input contains the question and selected evidence.

\begin{quote}
\small
\raggedright
Answer the question using only the evidence below.
Return the shortest answer supported by the evidence
and do not add an explanation.
If the evidence is incomplete, return the best supported
answer from the available evidence.

Question: \{query\}\\
Evidence: \{selected evidence\}\\
Answer:
\end{quote}

\subsection{Preprocessing Prompts}

\paragraph{TAM abstraction generator.}
After group membership is determined at a level,
the auxiliary model summarizes each group's final members.
The target node type determines whether the summary
describes a relation or a topic.

\begin{quote}
\small
\raggedright
You generate a concise textual abstraction for a typed
retrieval hierarchy.
Use only the information provided by the child nodes.

For a Relation node, describe the semantic relation
connecting the supplied evidence regions.
For a Topic node, describe the broader subject shared
by the supplied Relation nodes.
Preserve names, dates, entities, and relations that
may be needed for retrieval.
Do not add unsupported information and do not answer
the user query.

Node type: \{node type\}\\
Child nodes: \{child-node text\}

Output:\\
\texttt{ABSTRACTION:} \{summary\}
\end{quote}

\paragraph{DocVQA front end.}
The shared Qwen-VL-Plus front end converts document images
into textual regions before retrieval.
Page and region identifiers preserve the connection
between the extracted text and its document source.

\begin{quote}
\small
\raggedright
Convert the supplied document image into faithful textual
evidence for document question answering.
Preserve visible text, headings, labels, table entries,
and local associations needed to interpret the document.
Organize the output into concise textual regions
in reading order.

Do not answer the user query and do not add information
that is not visible in the document.
Associate each region with its page and region identifier.

Document image: \{document image\}

Output for each region:\\
\texttt{PAGE:} \{page identifier\}\\
\texttt{REGION\_ID:} \{region identifier\}\\
\texttt{TEXT:} \{faithful textual content\}
\end{quote}

\subsection{Full Revision Input}

When an additional update check rejects a recoverable
proposal, Full invokes the producing role with its original
task instruction and the following revision context.
The controller supplies the check feedback, available
candidates and references, and remaining budget.

\begin{quote}
\small
\raggedright
Current question:\\
\{question\}

Accepted state available to this role:\\
\{accepted state\}

Rejected proposal:\\
\{rejected proposal\}

Update-check feedback:\\
Target field: \{field\}\\
Error type: \{error code\}\\
Observed relation: \{observed relation\}\\
Required relation: \{required relation\}

Available candidates and references:\\
\{allowed candidates and references\}

Remaining budget:\\
\{remaining budget\}

Revise the proposal to address the reported error.
Use only the available sources and references.
Do not modify fields outside your assigned role.
Return the revised proposal using the role's required
output format.
\end{quote}

This block addresses update rejection.
An accepted \textsc{Fail} assessment instead supplies
evidence-gap feedback to the next retrieval plan.
Revision is performed by the original role, and the
resulting proposal passes through the same update checks.
All revision calls count toward the shared query-time
auxiliary-call budget.

\section{Backend and Memory Implementation}
\label{app:implementation}

\subsection{TAM Construction}
\label{app:tam_details}

TAM organizes the source collection into Root, Topic,
Relation, and Evidence nodes.
Documents are divided into overlapping regions.
Each region becomes an Evidence node that retains
its source identifier and offsets.
Region embeddings support cosine-based agglomerative
grouping into Relation groups, which are subsequently
grouped into Topic groups.
Table~\ref{tab:appendix_tam_settings} lists the construction
and navigation parameters.

\begin{table*}[t]
    \centering
    \small
    \setlength{\tabcolsep}{5pt}
    \renewcommand{\arraystretch}{1.12}
    \begin{tabularx}{\textwidth}{
        @{}p{0.30\textwidth}
        >{\raggedright\arraybackslash}X@{}
    }
        \toprule
        Item & Setting \\
        \midrule

        Region length
        & 384 tokens \\

        Region overlap
        & 64 tokens \\

        Embedding encoder
        & \texttt{BAAI/bge-base-en-v1.5} \\

        Base grouping
        & Cosine-based agglomerative grouping \\

        Relation grouping threshold
        & 0.78 \\

        Relation group capacity
        & At most eight Evidence nodes \\

        Topic grouping threshold
        & 0.70 \\

        Topic group capacity
        & At most six Relation nodes \\

        Cross-group compatibility
        & Node--group cosine similarity of at least
          0.78 for Evidence-to-Relation assignment
          and 0.70 for Relation-to-Topic assignment \\

        Parent limit
        & At most two parent groups per node \\

        Group representation
        & Normalized mean of member embeddings,
          fixed during cross-group assignment \\

        Abstraction generator
        & Qwen2.5-7B, temperature 0 \\

        Trace scorer
        & Batched Qwen2.5-7B scoring, temperature 0 \\

        Local branch limit
        & Two retained extensions per active trace \\

        Beam width
        & Four \\

        Candidate budget
        & At most 32 candidate extensions per traversal \\

        Maximum navigation depth
        & Three edges from the root \\

        Tie breaking
        & Stable node order \\

        Construction scope
        & The source collection associated with
          the current query \\

        \bottomrule
    \end{tabularx}
    \caption{TAM construction and navigation settings.
    Cross-group assignment also respects node types,
    adjacent levels, group capacities, and the parent limit.}
    \label{tab:appendix_tam_settings}
\end{table*}

\begin{table*}[!t]
    \centering
    \small
    \setlength{\tabcolsep}{5pt}
    \renewcommand{\arraystretch}{1.12}
    \begin{tabularx}{\textwidth}{
        @{}p{0.18\textwidth}
        >{\raggedright\arraybackslash}X
        >{\raggedright\arraybackslash}X@{}
    }
        \toprule
        Scope & Stored content & Access and lifetime \\
        \midrule

        Global $M_g$
        & Document chunks and embeddings;
          LLM-generated document summaries and
          Relation/Topic abstractions;
          accepted retrieval-location records
          with node references and source associations
        & Can persist across queries when enabled.
          Reused content must refer to source material
          accessible to the current query. \\

        Query-scoped $M_t$
        & Shared plans, retrieval candidates,
          selected evidence, and verification records
        & Available to permitted roles working on
          the current query.
          Initialized for each new query. \\

        Role-private $M_a$
        & The owning role's intermediate records,
          such as its planning history
        & Available only to its owning role within
          the current query.
          Initialized for each new query. \\

        \bottomrule
    \end{tabularx}
    \caption{SMP contents and memory scopes.
    The field $U$ holds references to stored results,
    while $E$ contains evidence selected for assessment
    and answering.}
    \label{tab:appendix_memory}
\end{table*}

\paragraph{Cross-group assignment.}
Base grouping first assigns each node to a parent group.
The normalized mean of the group's member embeddings
then serves as its representation.
These representations remain fixed while nodes are
considered for additional membership.

A node can acquire a second parent when its cosine
similarity to another group's representation meets
the corresponding threshold.
The assignment must also satisfy the node-type,
level, and group-capacity constraints.
The original parent edge is retained, and the additional
edge points to the same child-node identifier.
Each node has at most two parents.
New assignments do not change the group representations
used for compatibility scoring during that step.

\paragraph{Abstraction generation.}
After membership is finalized at a level, the auxiliary
model summarizes each group's members using the prompt
in Appendix~\ref{app:role_prompts}.
Relation summaries describe connections among evidence
regions. Topic summaries describe subjects shared by
Relation nodes.
A root connects the Topic nodes.
Edges run only from a parent level to the adjacent
child level, forming a directed acyclic graph.

Evidence nodes locate source regions.
The Retriever obtains passages from the selected region,
and the passages admitted for assessment and answering
are stored in the state field $E$.

\paragraph{Construction procedure.}
Algorithm~\ref{alg:tam_construction} constructs TAM from
Evidence regions to Relation and Topic abstractions.
\textsc{BaseGroups} performs cosine-based agglomerative grouping
with the specified threshold and capacity, initially assigning
each child to one group. Cross-group assignment uses fixed group
representations and adds at most one further parent per child.

\begin{algorithm}[t]
\caption{TAM construction with cross-group assignment}
\label{alg:tam_construction}
\footnotesize
\begin{algorithmic}[1]
\Require Source documents $D$, embedding encoder $f_{\mathrm{emb}}$,
         abstraction generator $F_a$
\Ensure Typed graph $H$ with Root, Topic, Relation,
        and Evidence nodes

\State $H \gets$ empty graph, $L \gets \emptyset$

\For{$x$ in $\Call{Segment}{D,384,64}$}
    \State Create Evidence node $v_x$ with the text,
        source identifier, and offsets of $x$
    \State $e(v_x) \gets f_{\mathrm{emb}}(\mathrm{text}(x))$
    \State Add $v_x$ to $H$ and $L$
\EndFor

\For{$(\ell,\tau,k)$ in
    $[(\mathrm{Relation},0.78,8),(\mathrm{Topic},0.70,6)]$}
    \State $\mathcal{G} \gets \Call{BaseGroups}{L,\tau,k}$

    \For{$G$ in $\mathcal{G}$}
        \State Create a parent node $v_G$ of type $\ell$
            and add it to $H$
        \State Add edges $(v_G,v)$ for every $v\in G$
        \State $\mu_G \gets
            \operatorname{Normalize}
            (\operatorname{Mean}\{e(v):v\in G\})$
    \EndFor

    \State Keep all $\mu_G$ fixed during the following
        assignment step

    \For{$v$ in $L$}
        \State $\mathcal{C}\gets
            \{G\in\mathcal{G}:v\notin G,\ |G|<k,\
            \cos(e(v),\mu_G)\geq\tau\}$
        \State Retain only groups in $\mathcal{C}$
            with a valid parent type and adjacent level

        \If{$\mathcal{C}\neq\emptyset$
            and $v$ has fewer than two parents}
            \State Select one compatible group
                $G^\star\in\mathcal{C}$
            \State Add $v$ to the member set of
                $G^\star$ in $\mathcal{G}$
            \State Add edge $(v_{G^\star},v)$ without changing
                $v$'s identifier or original parent edge
        \EndIf
    \EndFor

    \For{$G$ in $\mathcal{G}$}
        \State $a(v_G) \gets F_a(\ell,\{a(v):v\in G\})$
            \Comment{Summarize final members}
        \State $e(v_G) \gets f_{\mathrm{emb}}(a(v_G))$
    \EndFor

    \State $L\gets\{v_G:G\in\mathcal{G}\}$
\EndFor

\State Create Root node $v_{\mathrm{root}}$ and connect it
    to every Topic node in $L$
\State \Return $H$
\end{algorithmic}
\end{algorithm}

For Evidence nodes, $a(v)$ is the source-region text.
For Relation and Topic nodes, it is the generated abstraction.
Group capacity includes additional members, while compatibility
scores continue to use the representations computed from base
groups. All edges connect adjacent levels, preserving the DAG.

\subsection{TAM Navigation}

The Navigator starts at the root and expands paths
through admissible edges.
The scorer receives the question, current plan,
and candidate node sequences with their abstractions.
It returns a score for each candidate identifier.
For each active trace, at most two extensions are retained,
and the beam contains at most four active traces.
The traversal also respects the candidate and depth
limits in Table~\ref{tab:appendix_tam_settings}.

A completed trace ends at an Evidence node.
The Navigator selects a completed admissible trace
using the scores and constructs its proposal for $\pi$
from the corresponding candidate record.
The scorer does not generate the node sequence.
After the proposal is accepted, its terminal region
determines the candidates available to the Retriever.

\subsection{Flat Retrieval}
\label{app:flat_details}

The Flat backend replaces TAM traversal with passage
retrieval over the same segmented source material.
The Navigator submits a retrieval query and collects
ranked passage identifiers and scores.
It combines these results with the request and backend
metadata to construct $\pi$.
After this record is accepted, the Retriever resolves
passage identifiers to text and source information.
The retrieved passages enter the same evidence preparation,
assessment, and answer-generation workflow.
The record fields are listed in
Appendix~\ref{app:state_example}.

\subsection{SMP Contents and Access}
\label{app:smp_details}

SMP stores intermediate results for later lookup and reuse.
These results include document-processing outputs,
generated abstractions, retrieval-location records,
and records produced within the current query.
Table~\ref{tab:appendix_memory} summarizes the three
memory scopes.

\paragraph{Keys and references.}
Document-processing results are retrieved through
document cache keys.
Query-level records are associated with their function
and retrieval round, for example
\texttt{plan:r}, \texttt{candidates:r},
\texttt{evidence:r}, and \texttt{verification:r},
where $r$ denotes the retrieval round.
Role-private records also retain their owning role.

A memory request specifies the requesting role,
operation, scope, and key.
The access policy checks these attributes before
returning or storing content.
References in $U$ identify stored objects.
Their contents remain in SMP and are retrieved
under the applicable access policy.

\paragraph{Reuse during execution.}
Different stored objects support different operations.
Reusing document chunks and embeddings avoids repeating
the corresponding preprocessing.
Reusing generated summaries and node abstractions
can skip their auxiliary-model generation calls.
Accepted retrieval-location records provide source-linked
context for planning or candidate selection within
the current query's accessible material.

These operations affect different parts of execution.
Chunk and embedding reuse reduces preprocessing work.
Summary reuse can reduce recorded LLM token usage.
Retrieval-location reuse can change the planning context,
candidate selection, and subsequent evidence.
Selected evidence still follows the common update
and assessment workflow before answer generation.

The persistence and reset conditions used in the main
comparisons and sequential-memory analysis are specified
in Appendix~\ref{sec:additional_setup}.

\end{document}